# Areas of Attention for Image Captioning

Marco Pedersoli[1]    Thomas Lucas[2]    Cordelia Schmid[2]    Jakob Verbeek[2]

[1] École de technologie supérieure, Montréal, Canada
[2] Univ. Grenoble Alpes, Inria, CNRS, Grenoble INP, LJK, 38000 Grenoble, France

[1] Marco.Pedersoli@etsmtl.ca    [2] firstname.lastname@inria.fr

## Abstract

*We propose "Areas of Attention", a novel attention-based model for automatic image captioning. Our approach models the dependencies between image regions, caption words, and the state of an RNN language model, using three pairwise interactions. In contrast to previous attention-based approaches that associate image regions only to the RNN state, our method allows a direct association between caption words and image regions. During training these associations are inferred from image-level captions, akin to weakly-supervised object detector training. These associations help to improve captioning by localizing the corresponding regions during testing. We also propose and compare different ways of generating attention areas: CNN activation grids, object proposals, and spatial transformers nets applied in a convolutional fashion. Spatial transformers give the best results. They allow for image specific attention areas, and can be trained jointly with the rest of the network. Our attention mechanism and spatial transformer attention areas together yield state-of-the-art results on the MSCOCO dataset.*

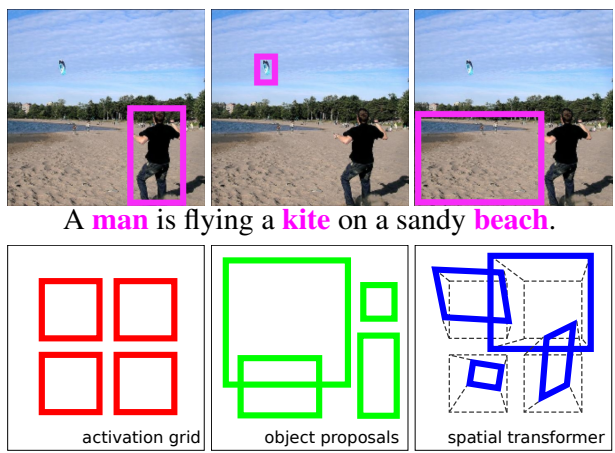

Figure 1. We propose an attention mechanism that jointly predicts the next caption word and the corresponding region at each time-step given the RNN state (top). Besides implementing our model using attention areas defined over CNN activation grids or object proposals, as used in previous work, we also present a end-to-end trainable convolutional spatial transformer approach to compute image specific attention areas (bottom).

## 1. Introduction

Image captioning, *i.e.* automatically generating natural language image descriptions, is useful for the visually impaired, and for natural language based image search. It is significantly more challenging than classic vision tasks such as object recognition and image classification for two reasons. First, the structured output space of well formed natural language sentences is significantly more challenging to predict over than just a set of class labels. Second, this complex output space allows a finer interpretation of the visual scene, and therefore also requires a more detailed visual analysis of the scene to do well at this task. Figure 1(top) gives an example of a typical image description that not only refers to objects in the scene, but also the scene type or location, object properties, and their interactions.

Neural encoder-decoder based approaches, similar to those used in machine translation [30], have been found very effective for this task, see *e.g.* [19, 23, 32]. These methods use a convolutional neural network (CNN) to encode the input image into a compact representation. A recurrent neural network (RNN) is used to decode this representation word-by-word into a natural language description of the image. While effective, these models are limited in that the image analysis is (i) static, *i.e.* does not change over time as the description is produced, and (ii) not spatially localized, *i.e.* describes the scene as a whole instead of focusing on local aspects relevant to parts of the description. Attention mechanisms can address these limitations by dynamically focusing on different parts of the input as the output sequence is generated. Such mechanisms are effective for a variety of sequential prediction tasks, including ma-chine translation [1], speech recognition [4], image synthesis [11], and image captioning [34]. For some tasks the definition of parts of the input to attend to are clear and limited in number: for example the individual words in the source sentence for machine translation. For other tasks with complex inputs, such as image captioning, the notion of parts is less clear. In this work we propose a novel attention model and three different ways to select parts of the image, or areas of attention, for the automatic generation of image captions.

The first contribution of our work is a new attention mechanism that models the interplay between the RNN state, image region descriptors, and word embedding vectors by means of three pairwise interactions. Previous attention approaches model either only the interaction between image regions and RNN state [15, 34], or the interaction between regions and words but with an external representation that is learned off-line, *e.g.* pre-trained object detectors [9, 33, 38]. In contrast, our attention representation explicitly considers, in a single end-to-end trainable system, the direct interaction among caption words, image regions and RNN state. At each time-step, our model jointly predicts the next caption word and the associated image region. Similar to weakly-supervised object localization, the associations between image regions and words are inferred during training from image-level captions. Our experimental results show that our three pair-wise interactions clearly improve the attention focus and the quality of the generated sentences.

Our second contribution is to integrate a localization sub-network in our model —similar to spatial transformer networks [14], but applied in a convolutional fashion— that regresses a set of attention areas from the image content. Earlier attention-based image captioning models used the positions in the activation grid of a CNN layer as attention areas, see *e.g.* [34]; such regions are not adaptive to the image content. Others have used object proposals as attention regions, see *e.g.* [15], in which case the regions are obtained by an external mechanism, such as Edge boxes [39], that is not trained jointly with the rest of the captioning system.

Our third contribution is a systematic experimental study of the effectiveness of these three different areas of attention using a common attention model, see Figure 1(bottom). To the best of our knowledge we are the first to present such a comparison. Our experimental results show that the use of image-specific areas of attention is important for improved sentence generation. In particular, our spatial-transformer based approach is a good choice: it outperforms the other approaches, while using fewer regions and not requiring an external proposal mechanism. Using our proposed attention mechanism and the spatial transformer attention areas together we obtain state-of-the-art performance on the MSCOCO dataset.

## 2. Related work

Image captioning with encoder-decoder models has recently been extensively studied, see *e.g.* [2, 8, 17, 19, 23, 25, 32, 34, 35]. In its basic form a CNN processes the input image to encode it into a vectorial representation, which is used as the initial input for an RNN. Given the previous word, the RNN sequentially predicts the next word in the caption without the need to restrict the temporal dependence to a fixed order, as in approaches based on n-grams. The CNN image representation can be entered into the RNN in different manners. While some authors [17, 32] use it only to compute the initial state of the RNN, others enter it in each RNN iteration [8, 23].

Xu *et al*. [34] were the first to propose an attention-based approach for image captioning, in which the RNN state update includes the visual representation of an image region. Which image region is attended to is determined based on the previous state of the RNN. They propose a "soft" variant in which a convex combination of different region descriptors is used, and a "hard" variant in which a single region is selected. The latter is found to perform slightly better, but is more complex to train due to a non-differentiable sampling operator in the state update. In thier approach the positions in the activation grid of a convolutional CNN layer is the loci of attention. Each position is described with the corresponding activation column across the layer's channels.

Several works build upon the approach of Xu *et al*. [34]. You *et al*. [38] learn a set of attribute detectors, similar to Fang *et al*. [9], for each word of their vocabulary. These detectors are applied to an image, and the strongest object detections are used as regions for an attention mechanism similar to that of Xu *et al*. [34]. In their work the detectors are learned prior and independently from the language model. Wu *et al*. [33] also learn attribute detectors but manually merge word tenses (*walking*, *walks*) and plural/singulars (*dog*, *dogs*) to reduce the set of attributes. Jin *et al*. [15] explore the use of selective search object proposals [31] as regions of attention. They resize the regions to a fixed size and use the VGG16 [29] penultimate layer to characterize them. Yang *et al*. [35] improve the attention based encoder-decoder model by adding a reviewer module that improves the representation passed to the decoder. They show improved results for various tasks, including image captioning. Yao *et al*. [36] use a temporal version of the same mechanism to adaptively aggregate visual representations across video frames per word for video captioning. Yeung *et al*. [37] use a similar temporal attention model for temporal action localization.

Visual grounding of natural language expressions is a related problem [17, 27], which can be seen as an extension of weakly supervised object localization [3, 6, 28]. The goal is to localize objects referred to by natural language descriptions, while only using image-level supervision. Since the goal in visual grounding and weakly supervised localization is precise localization, methods typically rely on object proposal regions which are specifically designed to align well with object boundaries [31, 39]. Instead of localizing a given textual description, our approach uses image-level supervision to infer a latent correspondence between the words in the caption and image regions.

Object proposal methods were designed to focus computation of object detectors on a selective set of image regions likely to contain objects. Recent state-of-the-art detectors, however, integrate the object proposal generation and recognition into a single network. This is computationally more efficient and leads to more accurate results [22, 26]. Johnson *et al*. [16] use similar ideas for the task of localized image captioning, which predicts semantically relevant image regions together with their descriptions. In each region, they generate descriptions with a basic non-attentive image captioning model similar to the one used by Vinyals *et al*. [32]. They train their model from a set of bounding-boxes with corresponding captions per image. In our work we do not exploit any bounding-box level supervision, we instead infer the latent associations between caption words and image regions. We propose a convolutional variant of the spatial transformer network of Jaderberg *et al*. [14], to place the attention areas in an image-adaptive manner. This module is trained in an integrated end-to-end manner with the rest of our captioning model.

Compared to previous attention models [15, 34, 35, 38], our attention mechanism, consisting of a single interaction layer, is less complex yet improves performance. Our approach models a joint distribution over image regions and caption words, generalizing weakly supervised localization methods and RNN language models. It includes a region-word interaction found in weakly supervised localization, as well as a word-state interaction found in RNN language models. In addition, our model includes a region-state interaction which forms a dynamic appearance-based salience mechanism. Our model naturally handles different types of attention regions (fixed grid, object proposals, and spatial transformers), and is applicable to all tasks where attention can model joint distributions between parts of the input data and output symbols. To the best of our knowledge, we propose the first trainable image-adaptive method to define attention regions, and present the first systematic comparison among different region types for attention-based image captioning in a single model.

## 3. Attention in encoder-decoder captioning

In Section 3.1 we describe our baseline encoder-decoder model. We extend this baseline in Section 3.2 with our attention mechanism in a way that abstracts away from the underlying region types. In Section 3.3 we show how we integrate regions based on CNN activation grids, object proposals, and spatial transformers networks in our model.

### 3.1. Baseline CNN-RNN encoder-decoder model

Our baseline encoder-decoder model uses a CNN to encode an image $I$ into a vectorial representation $\phi(I) \in \mathbb{R}^{d_I}$, which is extracted from a fully connected layer of the CNN. The image encoding $\phi(I)$ is used to initialize the state of an RNN language model. Let $h_t$ denote the RNN state vector at time $t$, then $h_0 = \theta_{hi}\phi(I)$, where $\theta_{hi} \in \mathbb{R}^{d_h \times d_I}$ linearly maps $\phi(I)$ to the RNN state space of dimension $d_h$.

The distribution over $w_t$, the word at time $t$, is given by a logistic regression model over the RNN state vector,

$$p(w_t|h_t) \propto \exp\left(w_t^\top W \theta_{wh} h_t\right), \quad (1)$$

where $w_t \in \{0,1\}^{n_w}$ is a 1-hot coding over the captioning vocabulary of $n_w$ words, $W$ is a matrix which contains word embedding vectors as rows, and $\theta_{wh}$ maps the word embedding space to the RNN state space. For sake of clarity, we omit the dependence on $I$ in Eq. (1) and below.

We use an RNN based on gated recurrent units (GRU) [5], which are simpler than LSTM units [13], while we found them to be at least as effective in preliminary experiments. Abstracting away from the GRU internal gating mechanism (see supplementary material), the state update function is given by a non-linear deterministic function

$$h_{t+1} = g(h_t, W^\top w_t). \quad (2)$$

The feedback of $w_t$ in the state update makes that $w_{t+1}$ recursively depends on both $\phi(I)$ and the entire sequence of words, $w_{1:t} = (w_1, \ldots, w_t)$, generated so far.

During training we minimize the sum of losses induced by pairs of images $I_m$ with corresponding captions $w_{1:l_m}$,

$$\sum_m L(I_m, w_{1:l_m}, \theta) = -\sum_m \sum_{t=1}^{l_m} \ln p(w_t|h_t, \theta), \quad (3)$$

where $\theta$ collectively denotes all parameters of the CNN and RNN component. This amounts to approximate maximum likelihood estimation, due to local minima in the loss.

Once the model is trained, captions for a new image can be generated by sequentially sampling $w_t \sim p(w_t|h_t)$, and updating the state $h_{t+1} = g(h_t, w_t)$. Since determining the maximum likelihood sequence is intractable, we resort to beam search if a single high-scoring caption is required.

### 3.2. Attention for prediction and feedback

In the baseline model the image is used only to initialize the RNN, assuming that the memory of the recurrent net is sufficient to retain the relevant information of the visual scene. We now extend the baseline model with a mechanism to attend to different image regions as the caption is generated word-by-word. Inspired by weakly supervised object localization methods, we score region-word pairs and aggregate these scores by marginalization to obtain a predictive distribution over the next word in the caption. The advantage is that this model allows words to be associated with specific image region appearances instead of global image representations, which leads to better generalization to recognize familiar scene elements in novel compositions. Importantly, we maintain the word-state interaction in Eq. (1) of the baseline model, to ensure temporal coherence in the generated word sequence by recursive conditioning on all previous words. Finally, a region-state interaction term allows the model to highlight and suppress image regions based on their appearance and the state, implementing a dynamic salience mechanism. See Figure 2 for a schematic illustration of our model.

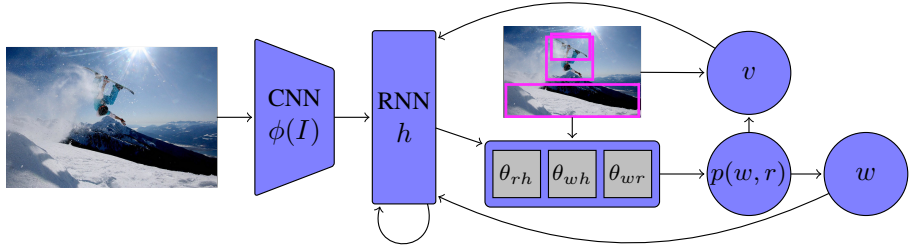

Figure 2. In our attention-based model the conditional joint distribution, $p(w, r|h)$, over words and regions given the current state $h$ is used to generate a word and to pool region descriptors in a convex combination. Both are then fed back to update the state at the next time-step.

We define a joint distribution, $p(w_t, r_t|h_t)$, over words $w_t$ and image regions $r_t$ at time $t$ given the RNN state $h_t$. The marginal distribution over words, $p(w_t|h_t)$, is used to predict the next word at every time-step, while the marginal distribution over regions, $p(r_t|h_t)$, is used to provide visual feedback to the RNN state update. Let $r_t \in \{0,1\}^{n_r}$ denote a 1-hot coding of the index of the region attended to among $n_r$ regions at time $t$. We write the state-conditional joint distribution on words and regions as

$$p(w_t, r_t|h_t) \propto \exp s(w_t, r_t, h_t), \quad (4)$$
$$s(w_t, r_t, h_t) = w_t^\top W \theta_{wh} h_t + w_t^\top W \theta_{wr} R^\top r_t$$
$$+ r_t^\top R \theta_{rh} h_t + w_t^\top W \theta_w + r_t^\top R \theta_r, \quad (5)$$

where $R$ contains the region descriptors in its rows. The score function $s(w_t, r_t, h_t)$ is composed of three bi-linear pairwise interactions. The first scores state-word combinations, as in the baseline model. The second scores the compatibility between words and region appearances, as in weakly supervised object localization. The third scores region appearances given the current state, and acts as a dynamic salience term. The last two unary terms implement linear bias terms for words and regions respectively.

Given the RNN state, the next word in the image caption is predicted using the marginal word distribution, $p(w_t|h_t) = \sum_{r_t} p(w_t, r_t|h_t)$, which replaces Eq. (1) of the baseline model. The baseline model is recovered for $R = 0$.

In addition to using the image regions to extend the state-conditional word prediction model, we also use them to extend the feedback connections of the RNN state update. We use a mechanism related to the soft attention model of Xu et al. [34]. We compute a convex combination of region descriptors which will enter into the state-update. In contrast to Xu et al., we derive the region weights from the joint distribution defined above. In particular, we use the marginal distribution over regions, $p(r_t|h_t) = \sum_{w_t} p(w_t, r_t|h_t)$, to pool the region descriptors as

$$v_t = \sum_{r_t} p(r_t|h_t) r_t^\top R = p_{rh}^\top R, \quad (6)$$

where $p_{rh} \in \mathbb{R}^{n_r}$ stacks all region probabilities at time $t$. This visual representation is concatenated to the generated word in the feedback signal of the state update, i.e. we replace the update of Eq. (2) of the baseline model with

$$h_{t+1} = g(h_t, [w_t^\top W \ v_t^\top]^\top). \quad (7)$$

In Section 4, we experimentally assess the importance of the different pairwise interactions, and the use of the attention mechanism in the state update.

### 3.3. Areas of attention

Our attention mechanism presented above is agnostic to the definition of the attention regions. In this section we describe how to integrate three types of regions in our model.

**Activation grid.** For the most basic notion of image regions we follow the approach of Xu et al. [34]. In this case the regions of attention correspond to the $z = x \times y$ spatial positions in the activation grid of a CNN layer $\gamma(I)$ with $c$ channels. The region descriptors in the rows of $R \in \mathbb{R}^{z \times c}$ are given by the activations corresponding to each one of the $z$ locations of the activation grid. In this case, the receptive fields for the regions is the same as all regions have a fixed shape and size, independent of the image content.

**Object proposals.** To obtain attention regions that adapt to the image content, we consider the use of object detection proposals, similar to the approach of Jin et al. [15]. We expect such regions to be more effective since they tend to focus on scene elements such as (groups of) objects, and their parts. In particular we use edge-boxes [39], and max-pool the activations in a CNN layer $\gamma(I)$ over each object proposal to obtain a set of fixed-size region descriptors. To ensure a high-enough resolution of the CNN layer which allows to pool activations for small proposals, we use a separate CNN which processes the input image at a higher resolution than the one used for the global image representation $\phi(I)$. This is similar to [10, 12], but we pool to a single cell instead of using a spatial pyramid. This is more efficient and did not deteriorate performance, as compared to using a pyramid. In this case the number of proposals is not limited by the number of positions in the activation tensor of the CNN layer that is accessed for the region descriptors.

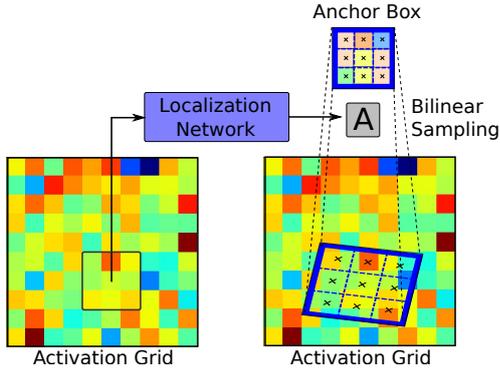

Figure 3. For our spatial transformer network attention areas, the localization network regresses affine transformations for all feature map positions in a convolutional manner, which are applied to the anchor boxes that are used to re-sample the feature map.

**Spatial transformers.** We propose a third type of attention region that has not been used in existing attention-based captioning models. It is inspired by recent object detectors and localized image captioning methods with integrated the region proposal networks [16, 22, 26]. In contrast to the latter methods, which rely on bounding-box annotations to learn the region proposal network, we only use image captions for training. Therefore, we need a mechanism that allows back-propagation of the gradient of the captioning loss w.r.t. the region coordinates and the features extracted using them. To this end we use a bilinear sampling approach as in [14, 16]. In contrast to the max-pooling we use for proposals, it enables differentiation w.r.t. the region coordinates.

Our approach is illustrated in Figure 3. Given an activation map $\gamma(I)$, we use a localization network that consists of two convolutional layers to locally regress an affine transformation $A \in \mathbb{R}^{2 \times 3}$ for each location of the feature map. With each location of the activation map $\gamma(I)$ we associate an "anchor box", which is centered at that position and covers $3 \times 3$ activations. The affine transformations, computed at each location in a convolutional fashion, are applied to the coordinates of the anchor boxes. Locally a $3 \times 3$ patch is bilinearly interpolated from $\gamma(I)$ over the area of the transformed anchor box. A $3 \times 3$ filter is then applied to the locally extracted patches to compute the region descriptor, which has the same number of dimensions as the activation tensor $\gamma(I)$ has channels. If the local transformations leave the anchor boxes unchanged, then this reduces to the activation grid approach.

As we have no bounding-box annotations, training the spatial transformer can get stuck at poor local minima. To alleviate this issue, we initialize the network with a model that was trained using activation grids. We initialize the transformation layers to produce affine transformations that scale the anchor boxes to twice their original size, to move away from the local optimum of the activation grid model.

## 4. Experimental evaluation

We define the experimental setup in Section 4.1, and present the experimental results in Section 4.2.

### 4.1. Experimental setup and implementation details

**Dataset and evaluation metrics.** For most of our experiments we use the MSCOCO dataset [20]. It consists of around 80K training images and 40K development images. Each image comes with five descriptive captions, see Figure 5 for example images. For sake of brevity we only report the most commonly used metrics, BLEU4, METEOR, and CIDEr-D, in the main paper. BLEU 1, 2 and 3 metrics can be found in the supplementary material. Similar to previous work [33, 34, 35] we use 5K development images to validate the training hyper-parameters based on CIDEr-D and another 5K development images to measure performance. Finally, we also use the visual entity annotations of Plummer *et al*. [24] to assess the extent to which the attention model focuses on objects or their context.

**CNN image encoder.** We use the penultimate layer of the VGG16 architecture [29] to extract the global image representation $\phi(I)$ that initializes the RNN state. The "activation grid" regions are taken from the last convolutional layer. For the "spatial transformer" regions, we use the penultimate convolutional layer to regress the transformations, which are then applied to convolve a locally transformed version of the same layer. For the "object proposal" regions we max-pool features from the last convolutional layer. Similar to [26], we re-scale the image so that the smaller image dimension is 300 pixels while keeping the original aspect-ratio. When fine-tuning we do not share the parameters of the two CNNs. In all cases, the dimension of the region descriptors is given by the number of channels in the corresponding CNN layer, *i.e.* $d_r = 512$.

**Captioning vocabulary.** We use all 6,325 unique words in the training captions that appear at least 10 times. Words that appear less frequently are replaced by a special OUT-OF-VOCABULARY token, and the end of the caption is marked with a special STOP token. The word embedding vectors of dimension $d_w = 512$ collected in the matrix $W$ are learned along with the RNN parameters.

| Method | B4 | Meteor | CIDEr |
| --- | --- | --- | --- |
| Baseline: $\theta_{wh}$ | 26.4 | 22.2 | 78.9 |
| Ours: $\theta_{wh}, \theta_{wr}$ | 28.0 | 22.9 | 83.6 |
| Ours: $\theta_{wh}, \theta_{wr}, \theta_{rh}$ | 28.4 | 23.3 | 85.5 |
| Ours: conditional feedback | 28.7 | **23.7** | 86.8 |
| Ours: full model | **28.8** | **23.7** | **87.4** |

Table 1. Evaluation of the baseline and our attention model using activation grid regions, including variants with certain components omitted, and word-conditional instead of marginal feedback.

**Training.** We use RNNs with a single layer of $d_h = 512$ GRU units. We found it useful to train our models in two stages. In the first stage, we use pre-trained CNN weights obtained from the ImageNet 2010 dataset [7]. In the second stage, we also update the CNN parameters. We use the Adam stochastic gradient descend algorithm [18]. To speed-up training, we sub-sample the $14 \times 14$ convolutional layers to $7 \times 7$ when using the activation grid and the spatial transformer regions. For proposal regions, each time we process an image we use 50 randomly selected regions.

### 4.2. Experimental results

In this section we assess the relative importance of different components of our model, the effectiveness of the different types of attention regions, and the effect of jointly fine-tuning the CNN and RNN components. Finally, we compare our results to the state of the art.

**Attention and visual feedback.** In Table 1 we progressively add components of our model to the baseline system. Here we use activation grid regions for our attention model. Adding all components improves the CIDEr score of the baseline, 78.9, by 8.5 points to 87.4. The baseline RNN uses only word-state interaction terms to predict the next word given the RNN state. Adding the word-region interaction term (second row) improves the CIDEr metric by 4.7 points to 83.6. This demonstrates the significance of localized visual input to the RNN. As in weakly-supervised object detection, the model learns to associate caption terms to local appearances. Adding the third pairwise interaction term between regions and the RNN state (third row) brings another improvement of 1.9 points to 85.5 CIDEr. This shows that the RNN is also able to implement a dynamic salience mechanism that favors certain regions over others at a given time-step by scoring the compatibility between the RNN state and the region appearance. Finally we add the visual feedback mechanism to our model (87.4, last row), which drives the CIDEr-D score further up by 1.9 points. We also experimented with a word-conditional version of the visual feedback mechanism (86.8, last but one row), which uses $p(r_t|w_t, h_t)$ instead of $p(r_t|h_t)$ to compute the visual feedback. Although this also improves the CIDEr-D score, as compared to not using visual feedback, it is less effective than using the marginal distribution weights. The visualizations in Figure 5 suggest that the reason for this is that the marginal distribution already tends to focus on a single semantically meaningful area.

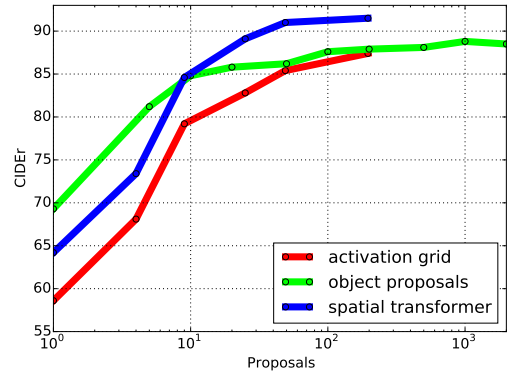

Figure 4. Image captioning performance in CIDEr-D as a function of the number of regions. Note the log-scale on the horizontal axis.

**Comparing areas of attention.** In our next set of experiments we compare the effectiveness of different attention regions in our model. In Figure 4 we consider the performance of the three region types as a function of the number of regions that are used when running the trained model on test images. For activation grids and spatial transformers the number of regions are regularly sampled from the original $14 \times 14$ resolution using increasing strides. For instance, using a stride of 2 generates $7 \times 7 = 49$ regions. For object proposals we test a larger range, from 1 up to 2,000 regions, sorted by their "objectness" score. For all three region types, performance quickly increases with the number of regions, and then plateaus off. Using four or less regions yields results below the baseline model, probably because strong sub-sampling at test-time is sub-optimal for models trained using $7 \times 7$ or 50 regions. The spatial transformer regions consistently improve over the activation grid ones, demonstrating the effectiveness of the region transformation sub-network. As compared to object proposals, the spatial transformer regions yield better results, while also being computationally more efficient: taking only 18ms to process an image using $7 \times 7$ regions, as compared to 352ms for 50 proposals which is dominated by 320ms needed to compute the proposals. At 6ms per image, fixed $7 \times 7$ activation grids are even more efficient, but come with less accurate results. In the remaining experiments, we report performance with the optimal number of regions per method: 1,000 for proposals, and 196 for grids and transformers.

**Joint CNN-RNN fine-tuning.** We now consider the effect of jointly fine-tuning the CNN and RNN components. In Table 2 we report the performance with and without fine-tuning for each region type, as well as the baseline performance for reference. All models are significantly improved by the fine-tuning. The baseline improves the most in ab-solute terms, but its performance remains substantially behind that of our attention models. The two types of image-dependent attention regions improve over fixed activation grids, but the differences between them are reduced after fine-tuning. Spatial transformer regions lead to comparable results as edge-box object proposals, that were designed to align with object boundaries. Spatial transformer regions, however, are more appealing from a modeling perspective since the region module is trainable fully end-to-end and does not rely on an external image processing pipeline, while also being more efficient to compute.

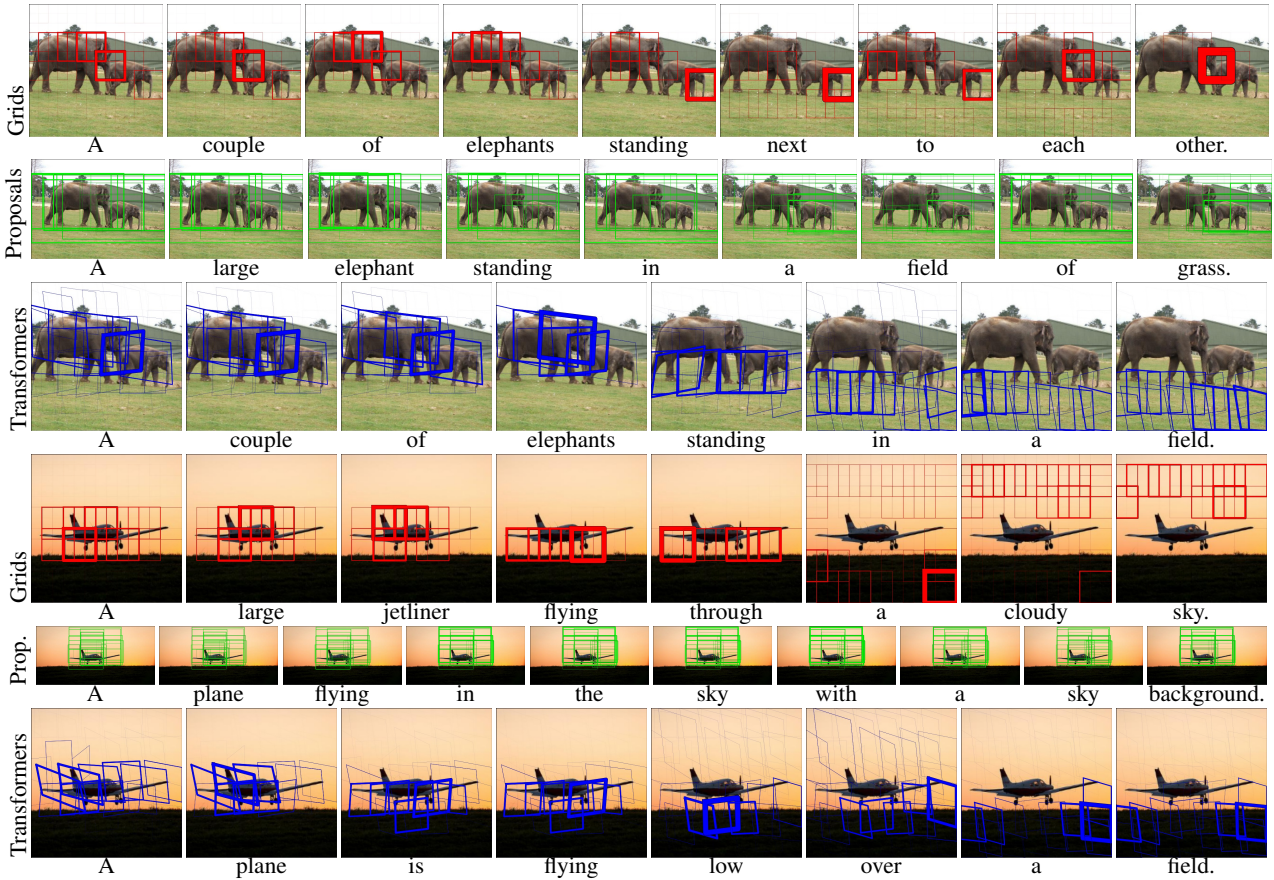

Figure 5. Visualization of the focus of our attention model during sequential word generation for the three different region types: activation grids, object proposals, and spatial transformers. The attention areas are drawn with line widths directly proportional to weights $p(r_t|h_t)$.

|  | B4 | Meteor | CIDEr |
|---|---|---|---|
| RNN training only | | | |
| Baseline | 26.4 | 22.2 | 78.9 |
| Activation grid | 28.8 | 23.6 | 87.4 |
| Object proposals | 28.9 | 23.7 | 89.0 |
| Spatial transformers | **30.2** | **24.2** | **91.1** |
| CNN-RNN fine-tuning | | | |
| Baseline | 28.7 | 23.5 | 87.1 |
| Activation grid | 30.3 | **24.5** | 92.6 |
| Object proposals | 30.1 | **24.5** | 93.7 |
| Spatial transformers | **30.7** | **24.5** | **93.8** |

Table 2. Captioning performance of the baseline and our model using different attention regions, with and without fine tuning.

**Visualizing areas of attention.** In Figure 5 we provide a qualitative comparison of the attentive focus using different regions in our model. A larger selection, including failure cases, can be found in the supplementary material. We show the attention weights over the image regions at each point in the generated sentences. For the spatial transformers, we show the transformed anchor boxes. For the activation grid regions, we show the back-projection of a $3 \times 3$ activation block, which allows for direct comparison with the spatial transformers. Note that in all cases the underlying receptive fields are significantly larger than the depicted areas. For object proposals we directly show the edge-box proposals. The images displayed for the object proposals differ slightly from the others, since the high-resolution network used in that case applies a different cropping and scaling scheme. Proposals accurately capture objects, e.g. the elephants and the plane, but in other cases regions for background elements are missing, e.g. for the field and the sky. The spatial transformers tend to focus quite well on relational terms. For example, *"standing"* focuses on the area around the legs of the elephants in the first image, and *"low"* on the area between the airplane and the ground in the second image. For the spatial transformers in particular, the focus of attention tends to be stable across meaningful sub-sequences, such as noun phrases (*e.g. "A couple of elephants"*) and verb phrases (*e.g. "is flying."*).

| Method | GT | Gen. |
|---|---|---|
| Liu *et al*. [21] | 38.4 | 52.0 |
| Liu *et al*. [21], spatial superv. | 43.3 | 57.9 |
| Areas of Attention, MSCOCO | 42.4 | 68.5 |
| Areas of Attention, Flickr30k | 40.2 | 61.1 |

Table 3. Attention correctness for ground truth (GT) and generated (Gen.) sentences on the Flickr30k test set.

**Attention correctness.** We follow the approach of Liu *et al*. [21] to quantitatively assess the alignment of attention with image regions corresponding to the generated caption words. Their approach uses the visual entity annotations on the Flickr30k dataset by Plummer *et al*. [24]. For caption words that are associated with a ground-truth image region, they integrate the attention values over that region. See Liu *et al*. [21] for more details. Following the protocol of Liu *et al*., we measured the attention correctness of our model (based on spatial transformer regions) on MSCOCO for ground truth and generated sentences. As Liu *et al*. reported results with a model trained on Flickr30k, for a fairer comparison, we have also trained a model on Flickr30k using the same hyper-parameters and architecture as for MSCOCO. In terms of caption generation the model obtained a CIDEr of 41.3 and a BLEU4 of 22.2. As shown in Table 3, when considering the correctness computed on the ground truth sentences, both our models perform better than Liu *et al*. using the attention model of Xu *et al*. [34], and come close to their model trained with additional spatial supervision. However, when evaluating the attention correctness on the generated sentences, our models perform significantly better than those in Liu *et al*., including those trained with spatial supervision.

**Comparison to the state of the art.** We compare our results obtained using the spatial transformer regions to the state of the art in Table 4; we refer to our method as "Areas of Attention". We obtain state-of-the-art results on par with Wu *et al*. [33]. They use a region-based high-level attribute representation instead of a global CNN image descriptor to condition the RNN language model. This approach is complementary to ours. For sake of comparability, we also ensemble our model and compare to ensemble results in the bottom part of Table 4. For our ensemble, we trained using 30K additional validation images on top of the 80K training images, and use a random horizontal flip of the images during training. We use the same 5K validation images and 5K images for reporting as in the other experiments. We obtain state-of-the-art results, on par with Bengio *et al*. [2]. They used "scheduled sampling", a modified RNN training algorithm that samples from the generated words during training. With standard training, as for our results, they report 95.7 CIDEr.

| | B4 | Meteor | CIDEr |
|---|---|---|---|
| Xu *et al*. [34], soft | 24.3 | 23.9 | — |
| Xu *et al*. [34], hard | 25.0 | 23.0 | — |
| Yang *et al*. [35] | 29.0 | 23.7 | 88.6 |
| Jin *et al*. [15] | 28.2 | 23.5 | 83.8 |
| Donahue *et al*. [8] | 30.0 | 24.2 | 89.6 |
| Bengio *et al*. [2] | 30.6 | 24.3 | 92.1 |
| Wu *et al*. [33] | 31 | 26 | 94 |
| Areas of Attention | 30.7 | 24.5 | 93.8 |
| Ensemble methods | | | |
| Vinyals *et al*. [32] | 27.7 | 23.7 | 85.5 |
| You *et al*. [38] | 30.4 | 24.3 | — |
| Bengio *et al*. [2] | 32.3 | 25.4 | 98.7 |
| Areas of Attention | 31.9 | 25.2 | 98.1 |

Table 4. Comparison of our results to the state of the art on the MSCOCO dataset.

## 5. Conclusion

In this paper we made three contributions. (i) We presented a novel attention-based model for image captioning. Our model builds upon the recent family of encoder-decoder models. It is based on a score function that consists of three pairwise interactions between the RNN state, image regions, and caption words. (ii) We presented a novel region proposal network to derive image-specific areas of attention for our captioning model. Our region proposal network is based on a convolutional variant of spatial transformer networks, and is trained without bounding-box supervision. (iii) We evaluated our model with three different region types based on CNN activation grids, object proposals, and our region proposal network. Our extensive experimental evaluation shows the importance of all our model components, as well as the importance of image-adaptive attention regions. This work is a first step towards weakly-supervised learning of objects and relations from captions, *i.e.* short sentences describing the content of an image. Future work will improve these associations for example by training object and relation detectors based on them. We release an open source Theano-Lasagne based implementation of our model: https://github.com/marcopede/AreasOfAttention

**Acknowledgment.** We thank NVIDIA for donating GPUs used in this research. This work was partially supported by the grants ERC Allegro, ANR-16-CE23-0006, and ANR-11-LABX-0025-01.